\documentclass[conference]{IEEEtran}
\IEEEoverridecommandlockouts

\usepackage{cite}
\usepackage{makeidx}
\usepackage{url}
\usepackage{graphicx}
\usepackage{enumitem}
\usepackage{amsmath}
\usepackage{amsfonts}
\usepackage{amssymb}
\usepackage{amsthm}
\usepackage{algorithmic}
\usepackage{booktabs}
\usepackage{tabularx}
\usepackage{makecell}
\usepackage{float}
\usepackage{mathtools}
\usepackage{caption}
\usepackage{subcaption}
\usepackage{adjustbox}
\usepackage{xcolor}

\usepackage{bbm}

\theoremstyle{definition}

\newcommand{\refig}[1]{Fig.~\ref{#1}}
\newcommand{\refalg}[1]{Alg.~\ref{#1}}

\newcommand{\refsec}[1]{Sec.~\ref{#1}}

\usepackage[ruled,vlined,linesnumbered]{algorithm2e}
\SetAlgorithmName{Algorithm}{Algorithm}{List of Algorithms}

\DeclareMathOperator*{\argmin}{arg\,min}

\usepackage{makecell}
\usepackage{multirow}

\begin{document}

\title{SCoTT: Strategic Chain-of-Thought Tasking for Wireless-Aware Robot Navigation in Digital Twins
\thanks{
\noindent 
The TUM group acknowledges the support of the German BMFTR under the program “Souver\"an. Digital. Vernetzt.” as part of the research hubs 6G-life (Grant 16KISK002), QD-CamNetz (Grant 16KISQ077), QuaPhySI (Grant 16KIS1598K), and QUIET (Grant 16KISQ093). W. Saad was supported by the U.S. National Science Foundation under Grant CNS-222551.
}
}

\author{%
    \IEEEauthorblockN{%
        Aladin Djuhera\IEEEauthorrefmark{1},
        Amin Seffo\IEEEauthorrefmark{1},
        Vlad~C.~Andrei\IEEEauthorrefmark{1},
        Holger~Boche\IEEEauthorrefmark{1},
        Walid~Saad\IEEEauthorrefmark{2} \\
    }
    \IEEEauthorblockA{
        \IEEEauthorrefmark{1}Technical University of Munich, Munich, Germany,
        \IEEEauthorrefmark{2}Virginia Tech, Arlington, VA, USA\\
        Emails: \{aladin.djuhera, amin.seffo, vlad.andrei, boche\}@tum.de, walids@vt.edu
    }
}

\maketitle

% Abstract
\begin{abstract}
    Path planning under wireless performance constraints is a complex challenge in robot navigation. 
    However, naively incorporating such constraints into classical planning algorithms often incurs prohibitive search costs.
    In this paper, we propose SCoTT, a wireless-aware path planning framework that leverages vision-language models (VLMs) to co-optimize average path gains and trajectory length using wireless heatmap images and ray tracing data from a digital twin (DT). 
    At the core of our framework is Strategic Chain-of-Thought Tasking (SCoTT), a novel prompting paradigm that decomposes the exhaustive search problem into structured subtasks, each solved via chain-of-thought prompting.
    To establish strong baselines, we compare classical A* and wireless-aware extensions of it, and derive DP-WA*, an optimal, iterative dynamic programming algorithm that incorporates all path gains and distance metrics from the DT, but at significant computational cost.
    In extensive experiments, we show that SCoTT achieves path gains within 2\% of DP-WA* while consistently generating shorter trajectories.
    Moreover, SCoTT’s intermediate outputs can be used to accelerate DP-WA* by reducing its search space, saving up to 62\% in execution time.
    We validate our framework using four VLMs, demonstrating effectiveness across both large and small models, thus making it applicable to a wide range of compact models at low inference cost.
    We also show the practical viability of our approach by deploying SCoTT as a ROS node within Gazebo simulations.
    Finally, we discuss data-acquisition pipelines, compute requirements, and deployment considerations for VLMs in 6G-enabled DTs, underscoring the potential of natural language interfaces for wireless-aware navigation in real-world applications.
\end{abstract}

% Keywords

\IEEEpeerreviewmaketitle

% Sections

\section{Introduction and Motivation}
\label{sec:intro}

% 6G and DTs
Future 6G networks will underpin a host of emerging technologies, including artificial general intelligence (AGI) \cite{saad2024artificial}, autonomous transportation \cite{autonomous_transportation}, and the metaverse \cite{saad_wireless_metaverse}. 
At the same time, 6G-enabled digital twins (DTs) will be pivotal for smart manufacturing, intelligent planning, and autonomous robotics \cite{6G_digital_twin_saad}. 
DTs act as high-fidelity replicas of real physical systems and processes, thus encompassing not only their physical geometry but also their associated algorithms, communication protocols, and compute resources \cite{6G_digital_twin_saad}.
Embedding wireless ray tracing data into DTs will become critical for capturing realistic network-layer behavior and enabling connectivity-aware decision making.
In our prior work \cite{andrei2024digitaltwinningplatformintegrated}, for example, we developed a ray tracing-enabled DT framework for integrated sensing and communication (ISAC)-enabled robotics that fuses high-resolution wireless data with 3D environment models.
This allows for real-time, wireless-aware interactions between robots and their digital environments.

% DTs and LLMs
In future deployments, interacting with DTs through natural language will become increasingly important, particularly in human-in-the-loop systems \cite{llm_DT}, where operators can prompt a DT using a large language model (LLM) and, in effect, \emph{talk to the simulation}.
This shift raises a key question: How can we leverage the reasoning capabilities of LLMs within real-time DTs to support wireless-aware decision making?
In this paper, we consider the concrete use case where the LLM assistant is instructed to plan a wireless-aware path for robot navigation, e.g.:
\emph{``Please navigate from A to B using the shortest path possible while maintaining an average path gain of 10~dB.''}

% Path planning with LLMs
This example illustrates an inherently complex problem: Path planning with side constraints beyond simple distance minimization. 
In general, path planning is an exhaustive search problem and traditional algorithms become inefficient when operating over large datasets or incorporating complex side constraints.
This motivates research on whether path planning can be automated via LLMs. 
For example, S2RCQL \cite{S2RCQL} introduces a curriculum-learning approach with spatial-to-relational prompting, training the LLM to imitate reinforcement learning strategies. However, it struggles with scalability and often hallucinates in complex environments.
Similarly, LLM-A* \cite{llm_astar} augments the classical A* algorithm \cite{astar} by guiding the search with LLM-generated waypoints via few-shot prompting, reducing unnecessary exploration and increasing efficiency.

% SCoTT
However, reducing redundancy in wireless measurement data, such as path gains, may not be straightforward, posing a problem for data-efficient LLM-based path planning and requiring new approaches to how wireless data is represented, interpreted, and incorporated into the LLM.
In this work, we address this gap by introducing \emph{Strategic Chain-of-Thought Tasking} (SCoTT), a novel prompting paradigm that decomposes the wireless-aware path planning problem into structured subproblems, each solved through strategic chain-of-thought (SCoT) reasoning \cite{SCoT}, an extension of the popular chain-of-thought (CoT) prompting approach \cite{CoT}.
Specifically, we employ vision-language models (VLMs) and leverage retrieval-augmented generation (RAG) \cite{RAG} to process multi-modal wireless ray tracing data (wireless heatmap images and path gain data), generated by a real-world DT implementation from \cite{andrei2024digitaltwinningplatformintegrated}.
To demonstrate effectiveness, we apply SCoTT across four VLMs of varying sizes (Llama-4-Scout-17B \cite{llama4}, Llama-3.2-11B-Vision \cite{llama3}, SmolVLM \cite{smolvlm}, and Granite-Vision-3.2-2B \cite{granite}) and compare against wireless-aware extensions of the classical A* algorithm.
In summary, our key contributions are:
\begin{itemize}[leftmargin=*]
    \item 
    We first derive two wireless-aware extensions of A*:  
    (1) \emph{N-WA*} adds the inverse wireless path gain to the movement cost as a \emph{naive} baseline,
    and (2) \emph{DP-WA*} is an optimal, iterative \emph{dynamic programming} approach that evaluates all possible paths and jointly enforces both wireless and distance constraints using a Bellman-based cost-to-go.
    
    \item 
    Next, we introduce SCoTT, a novel prompting paradigm that improves language model-based path planning by decomposing the task into structured subtasks. 
    SCoTT requires the model to explain its decisions and uses multi-modal inputs (e.g., wireless heatmaps and path gain data) from a DT.
    In a structured three-stage process, SCoTT first generates a coarse path using vision prompting, then reduces the search space around this path, and finally computes a fine-grained solution using accurate path gain data from the DT.  
    
    \item 
    We conduct extensive simulations on three path planning examples using the four VLMs, comparing SCoTT to classical A*, N-WA*, and DP-WA*, and reporting model runtimes.
    Our results show that SCoTT achieves comparable performance to the cost-optimal DP-WA* while yielding shorter path lengths.
    We also demonstrate that DP-WA* can be significantly accelerated (by up to 62\%) when seeded with intermediate results from SCoTT’s first two stages.

    \item 
    To ensure real-world viability, we detail our robot operating system (ROS)-based integration of SCoTT within Gazebo \cite{gazebo} and present wireless-aware navigation outcomes in scenarios involving obstacle avoidance and motion control.

    \item 
    Finally, we discuss practical aspects of SCoTT, including data pipelines, compute requirements, and model deployment in 6G-enabled DTs. 
\end{itemize}

The remainder of the paper is organized as follows: 
\refsec{sec:wireless_ware_algorithmic_path_planning} introduces wireless-aware extensions of A*. 
\refsec{sec:SCoTT} presents our SCoTT framework. 
Simulation results are shown in \refsec{sec:simulation_results}, followed by the practical ROS integration in \refsec{sec:system_integration_and_practical_considerations} and the conclusions in \refsec{sec:conclusion}.
\section{Wireless-Aware Algorithmic Path Planning}
\label{sec:wireless_ware_algorithmic_path_planning}

\begin{figure}[!t]
    \centering
    \includegraphics[width=0.9\linewidth]{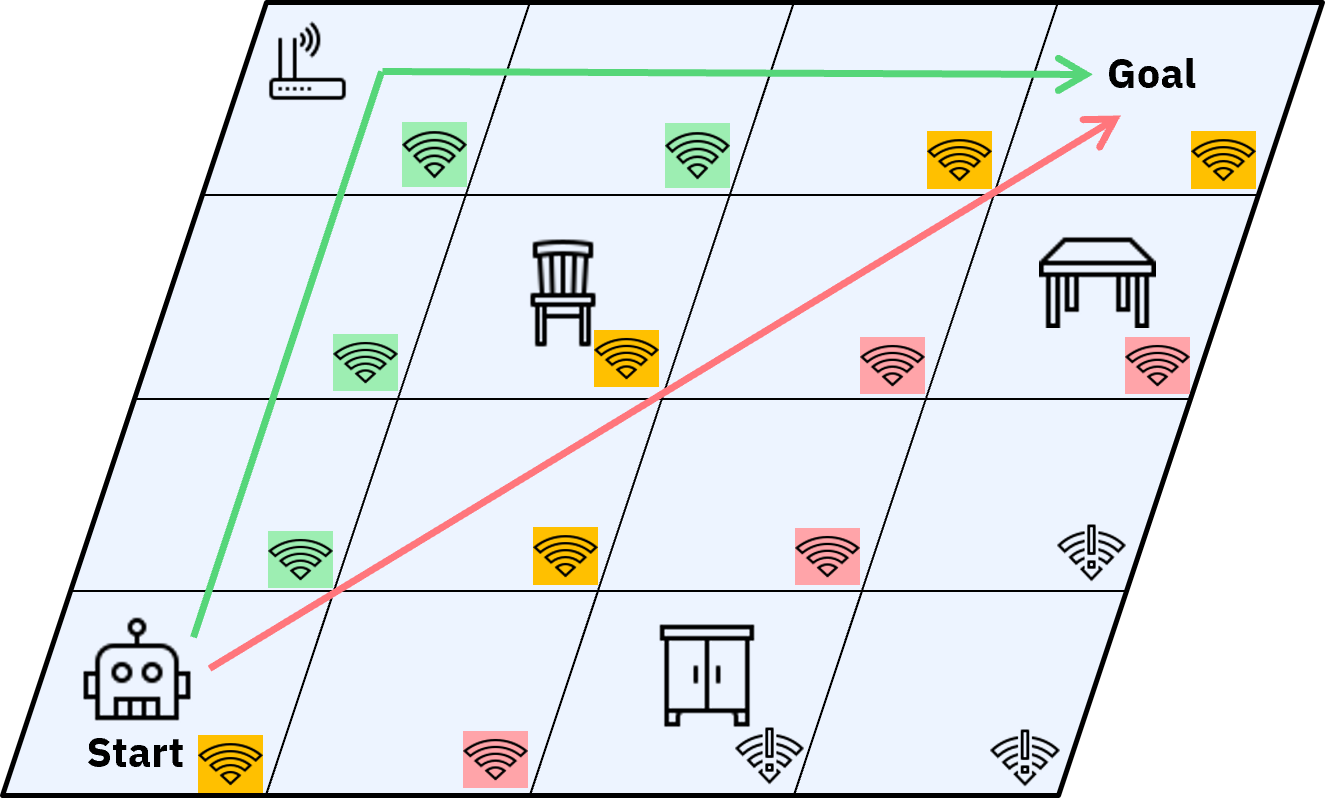}
    \caption{Wireless‐aware path planning objective: the green path is longer but offers better wireless coverage compared to the shorter red path.}
    \vspace{-0.5cm}
    \label{fig:path_planning_objective}
\end{figure}

In this paper, we focus on offline path planning with side constraints. 
A single-agent robot navigates from a start $n_{\text{start}}$ to a goal node $n_{\text{goal}}$ by choosing the shortest valid trajectory while ensuring that the average wireless path gain remains greater than some threshold $G$ (see \refig{fig:path_planning_objective}). 
We assume the DT environment to be fully-observable, static, and deterministic.

\subsection{N-WA*: Naive Wireless-Aware A*}
The classical A* algorithm finds the shortest path from $n_{\text{start}}$ to $n_{\text{goal}}$ by maintaining a balance between minimizing the path cost and reaching the goal efficiently. 
It is defined over the total cost function $f(n) = g(n) + h(n)$, where $g(n)$ represents the cost to reach node $n$ from $n_{\text{start}}$ and where $h(n)$ is a heuristic for the cost to reach $n_{\text{goal}}$ from node $n$. 
For simplicity, we assume a Euclidean cost environment. 
The algorithm iteratively selects the next node with the lowest $f(n)$ and updates the path cost and heuristic to find the optimal route. 
To incorporate wireless path gains, a simple though naive approach is to adapt the cost component by adding the inverse of the path gain value $p(n)$ at each node, i.e., 
\begin{align}
    g_{\text{N-WA*}}(n) = g(n) + \frac{1}{p(n) + \epsilon}, 
\end{align}
where the second term represents the additional cost-to-go with some small $\epsilon > 0$ to avoid division by zero. 
The updated total cost function is then $f_{\text{N-WA*}}(n) = g_{\text{N-WA*}}(n) + h(n)$. 

This approach is naive because even though it prefers nodes with higher path gains, it does not explicitly enforce any thresholds, making it a heuristic-based adaptation rather than a strict constraint optimization. Thus, N-WA* inherently biases toward the shortest path. 
This may lead to suboptimal paths when a high-gain route is longer, but ultimately preferable. 
Additionally, it cannot explicitly handle scenarios where an average path gain threshold $G$ must be maintained along the entire path. 
However, the additional storage of path gain values does not fundamentally alter the asymptotic complexity of the algorithm such that it has identical complexity as A*, i.e. $O(b^d)$, where $b$ is the average number of neighbors per node and $d$ is the number of steps in the optimal path. 
This makes N-WA* still inefficient for large data points, but avoids the substantial time and memory overhead of dynamic-programming (DP) solutions, as shown in the next section.

\subsection{DP-WA*: Dynamic Programming Wireless-Aware A*}
To move beyond heuristic-based planning and to achieve an effective trade-off between distance and average wireless path gains, we derive a multi-objective optimization problem using DP \cite{bellman2015applied}.
The problem is formulated on a rectangular grid representing the traversable area of the DT environment, where each point \( p_k \) at time step \( k \) corresponds to a specific discretized $(x, y)$ coordinate. 
The state transition model is given as \( p_{k+1} = f(p_k, u_k) \) and encodes movement in eight possible directions (north, north-west, etc.). 
The objective is to minimize the total travel distance while ensuring that the average wireless path gain remains above a threshold $G$. 
We formalize this multi-objective problem as follows:

\begin{align}
\min_{\{p_k\}_{k=0}^{T}} & \sum_{k=0}^{T-1} c(p_k, p_{k+1}) \quad \text{s.t.} \quad \frac{1}{T+1} \sum_{k=0}^{T} g(p_k) \geq G , \\
& p_{k+1} = f(p_k, u_k), \quad u_k \in \mathcal{U}(p_k), \quad \forall k , \\
& p_0 = \text{start position}, \quad p_T = \text{goal position} ,
\end{align}
where
\begin{itemize}[leftmargin=*]
    \item 
    \( c(p_k, p_{k+1}) \) is the cost of moving from position \( p_k \) to \( p_{k+1} \), in this case the Euclidean distance,
    
    \item 
    \( g(p_k) \) is the normalized path gain at position \( p_k \), provided by the DT and discretized between 0 and 1 in steps of 0.1,
    
    \item 
    \( u_k \) is an action out of the set of feasible actions \( \mathcal{U}(p_k) \).
\end{itemize}

Here, the state is the current position $p_k$ and $W_k$ are the accumulated path gains. 
We further let the value function $V_k(p_k, W_k)$ denote the minimal total distance from $p_k$ to $p_T$, given the accumulated $W_k$, i.e.,
\begin{align}
    V_k(p_k, W_k) = \min_{u_k \in \mathcal{U}(p_k)} \left\{ c(p_k, p_{k+1}) + V_{k+1}(p_{k+1}, W_{k+1}) \right\} , 
\end{align}
where $W_{k+1} = W_k + g(p_{k+1})$ is the new accumulated path gain. 
At the final time step $T$, we set the value function to
\begin{align}
    V_T(p_T, W_T) =
        \begin{cases}
        0, & \text{if } \dfrac{W_T}{T+1} \geq G \\
        \infty, & \text{otherwise}
        \end{cases} .
        \label{eq:boundary_condition}
\end{align}
This boundary condition ensures that only paths meeting the average wireless path gain threshold $G$ are considered feasible.

The corresponding \textit{recursive} Bellman equation then represents the minimal cost-to-go from $p_k$ to $p_T$, subject to maintaining the average path gain constraint, i.e.,
\begin{align}
    V_k(p_k, W_k) = & \min_{u_k \in \mathcal{U}(p_k)} \left\{ c(p_k, p_{k+1}) + V_{k+1}(p_{k+1}, W_{k+1}) \right\} \\
    & \text{s.t.} \quad W_{k+1} = W_k + g(p_{k+1}) 
\end{align}
with the feasibility condition
\begin{align}
    W_{k+1} + (T - (k+1)) \cdot g_{\max} \geq (T+1) \cdot G \ ,
    \label{eq:pruning_rule}
\end{align}
where $g_{\max} = 1$ is the maximum path gain at any position. 
This feasibility condition prunes states that cannot possibly meet the wireless path gain constraint by the final time step. 
With that, the optimal control policy at step $k$ selects the action $u_k^*$ that minimizes the total cost-to-go from $p_k$ under the wireless path gain constraint and is given by
\begin{align}
    u_k^* = \argmin_{u_k \in \mathcal{U}(p_k)} \left\{ c(p_k, p_{k+1}) + V_{k+1}(p_{k+1}, W_{k+1}) \right\} .
\end{align}

However, the recursive nature of this problem can lead to inefficiencies due to deep recursion and redundant recalculations. 
Instead, we use an \textit{iterative DP approach} where we fill in a table with intermediate results and compute the solution step-by-step by iteratively evaluating the value function \( V_k(p_k, W_k) \) in a bottom-up manner, starting backward in time from the final to the initial state. 
The concrete steps are outlined in \refalg{alg:DP-WA*}. After completing the iterations, the optimal path is reconstructed by starting from the initial state \( (p_0, W_0) \) and following the stored optimal actions \( u_k^* \) at each time step.

\begin{algorithm}[t]
    \footnotesize
    \caption{DP-WA* Algorithm}
    \label{alg:DP-WA*}

    % \vspace{1mm} \hrule \vspace{1mm}

    \textbf{Input:} Start position $p_0$, goal position $p_T$, grid $\mathcal{P}$ with wireless path gains $g(p)$, path gain threshold $G$, feasible actions $\mathcal{U}(p)$ \\
    \textbf{Output:} Optimal path from $p_0$ to $p_T$ satisfying $G$

    \vspace{1mm} \hrule \vspace{1mm}

    Initialize the DP table $V(p, W) \gets \infty$ for all positions $p$ and accumulated gains $W$ \\
    Set boundary condition at the goal according to \eqref{eq:boundary_condition}

    \For{$k = T-1$ down to $0$}{
        \For{each position $p$ in $\mathcal{P}$}{
            \For{each feasible action $u$ in $\mathcal{U}(p)$}{
                Compute next position $p' \gets f(p, u)$ \\
                Update accumulated gain $W' \gets W + g(p')$ \\
                \If{$W'$ satisfies the feasibility condition}{
                    Compute cost $C \gets c(p, p') + V(p', W')$ \\
                    \If{$C < V(p, W)$}{
                        Update $V(p, W) \gets C$ \\
                        Store optimal action $u^*(p) \gets u$ \\
                    }
                }
            }
        }
    }

    \If{$V(p_0, W_0) < \infty$}{
        Reconstruct the optimal path from $p_0$ using stored actions $u^*(p)$ \\
        \Return Optimal path
    }
    \Else{
        \Return No valid path exists
    }
\end{algorithm}

\begin{figure*}[htbp]
    \centering
    \includegraphics[width=1\textwidth]{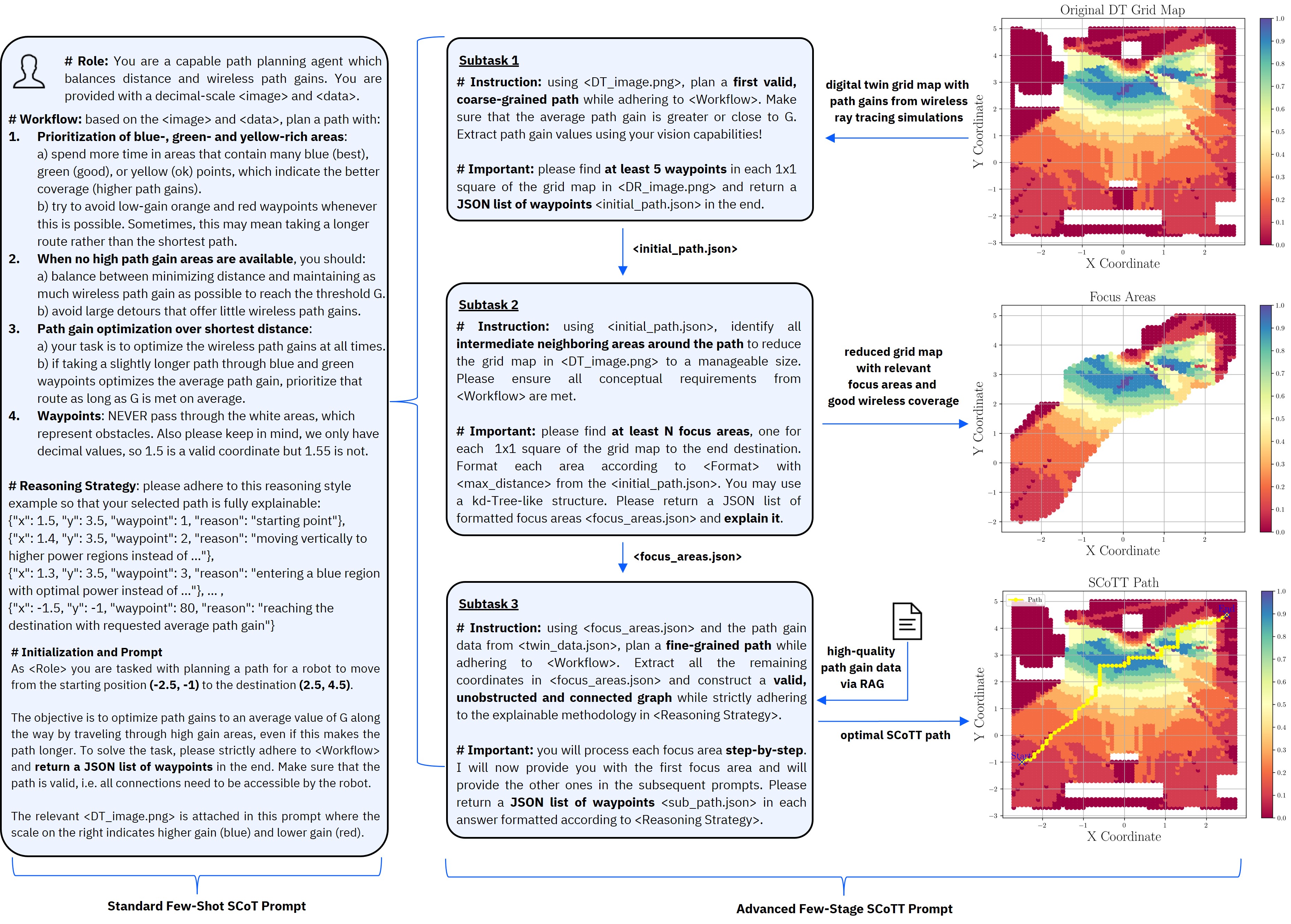}
    \caption{SCoTT prompt template that divides the complex wireless-aware path planning problem into three subtasks, each solved using strategic CoT prompting.}
    \label{fig:SCoTT_prompt}
\end{figure*}

This iterative realization of DP-WA* accurately models the trade-off between total travel distance and average path gains, implementing Bellman’s principle of optimality over a finite state–action graph. 
At each time \(k\), the value function $V_k(p_k, W_k)$ correctly captures the minimum remaining distance from \((p_k,W_k)\) to the goal, subject to the average path gain constraint $G$, ensured via our boundary condition in \eqref{eq:boundary_condition} and pruning rule in \eqref{eq:pruning_rule}. 
Since we systematically fill in every feasible \((p,W)\) pair from \(k=T\) back to \(k=0\), no potentially optimal path is ever discarded and DP-WA* \emph{always} finds the cost-optimal trajectory that meets the path gain threshold $G$.
However, despite avoiding redundant subproblem calculations, DP-WA* still incurs prohibitive time and space costs on large grids. 
Let the DT grid be \(N\times N\) so that \(\lvert\mathcal P\rvert = N^2\), and let the planning horizon \(T\) (i.e., number of steps) be at most proportional to \(N\) in typical traversals.
We further discretize accumulated path gains in integer increments of \(1/(T+1)\), such that there are at most \(T+1=O(N)\) distinct path gain levels \(W\).  
Thus, the total number of DP states is
\begin{align}
    \underbrace{\lvert\mathcal P\rvert}_{N^2}\;\times\;\underbrace{(T+1)}_{O(N)} \;=\; O(N^3).
\end{align}

Consequently, as each state evaluates a constant number of neighbor actions (eight in our 8-connected DT grid), both time and memory scale as \(O(N^3)\).
In contrast, N-WA* scales as \(O(b^d)\) with exponential dependence on path length \(d\), avoiding cubic growth in time and space but sacrificing the trade-off between distance and average path gain.  
While DP-WA* guarantees the cost-optimal path under the path gain constraint, its \(O(N^3)\) complexity renders it impractical on large grids without additional state reduction or coarser discretization.
\section{SCoTT: Stragetic Chain-of-Thought Tasking}
\label{sec:SCoTT}

SCoTT leverages multi-modal VLMs to enable efficient, wireless-aware path planning by targeting two core challenges of language model-based planners:
\begin{enumerate}
    \item \textbf{Hallucinations}: LLMs often fabricate waypoints or reasoning steps when asked to perform \emph{zero‐shot} planning with large data points (e.g., S2RCQL \cite{S2RCQL}, LLM-A* \cite{llm_astar}).

    \item \textbf{Scalability}: Limited context windows constrain the amount of data that can be processed in a single prompt, which makes it difficult to integrate large amounts of high-quality measurement data, e.g., from a DT.
\end{enumerate}

To overcome these limitations, SCoTT \emph{tasks} the VLM via a chain of smaller, strategy-driven prompts, thereby decomposing the complex wireless-aware path planning problem into focused subproblems, each solved independently via strategic chain-of-thought (SCoT) prompting \cite{SCoT}.
Unlike standard CoT \cite{CoT}, SCoT first elicits a high-level strategy for each subproblem and then uses that strategy to guide a sequence of reasoning steps, turning the monolithic task into a series of discrete, strategy-driven CoT tasks. 
In our wireless-aware path planning example, SCoT prompting involves two steps:
\begin{itemize}[leftmargin=*]
  \item \emph{Strategy Elicitation:} The model first outputs a concise strategy for the subproblem (e.g., ``identify high-gain corridors”).  
  \item \emph{Guided Reasoning:} The model then executes a short chain-of-thought following that strategy, focusing only on the relevant subset of the wireless path gain data from the DT.
\end{itemize}
By chaining these SCoT subtasks together, we realize \emph{Strategic Chain-of-Thought Tasking} (hence SCoTT), which avoids hallucinations (as each step is scoped and grounded) and context overflow (as each prompt handles a small slice of the problem).
\refig{fig:SCoTT_prompt} illustrates how SCoTT converts a single, context-limited few-shot SCoT prompt into a sequence of focused, few-stage subtasks.
Our prompt template carefully defines both the overall workflow (i.e., what needs to be achieved) and the reasoning strategy (i.e., how it needs to be done), ensuring the VLM remains grounded, understands limitations and boundaries, and explains each of its decisions, effectively overcoming hallucinations and context window constraints.
To this end, we provide the VLM with a multi-modal input:
\begin{itemize}
  \item \textbf{Image}: a bird’s‐eye‐view of the DT grid map, overlaid with a color-coded wireless path gain heatmap.
  \item \textbf{Text}: a JSON file of coordinates and precise path gain measurements, infused to the model via RAG. 
\end{itemize}

We decompose the planning task into consecutive SCoT prompts as follows. 
First, the VLM analyzes the wireless heatmap to identify high‐gain corridors of constraint-compliant waypoint regions. 
Then it performs three subtasks: 
(i) \emph{coarse path selection}, which generates an initial route, 
(ii) \emph{focus‐area definition}, which selects reduced search regions around the initial route, and 
(iii) \emph{fine‐grained waypoint refinement}, which computes the optimized, fine-grained path.
As shown in \refig{fig:SCoTT_prompt}, we prompt the VLM in three stages:

\begin{enumerate}[label=\textbf{Subtask \arabic*:},leftmargin=*]
  \item \emph{Coarse Path Strategy.} 
  We first prompt the VLM to find a coarse-grained initial path using the heatmap image while demanding a balance between good wireless coverage and distance via \textless Workflow\textgreater. 
  Therein, we specifically instruct the model to look for high gain paths (color-coded) and we define forbidden zones (e.g., obstacles in white) that must be avoided.
  This yields a valid, but not yet optimized first path.
  
  \item \emph{Focus‐Area Definition.} 
  Given the coarse path, we instruct the VLM to select \(N\) focus areas, each stored as a $k$d‐tree with radius \textless max\_distance\textgreater, serving as reduced search spaces around the initial route.
  Users may further adjust \(N\) and \textless max\_distance\textgreater \ to increase the number of focus areas, effectively broadening or narrowing the fine-grained search in the next step.

  \item \emph{Fine‐Grained Refinement.} 
  Finally, for each focus area, the model incrementally generates a detailed path using RAG-supplied ray tracing data, ensuring the wireless path gain threshold \(G\) is met. 
  Because each prompt now covers only a small region, the context window easily holds both instructions and data, allowing vision-guided waypoint selection with precise gain constraints.  
  By iterating over these compact areas, the model avoids large-scale hallucinations and consistently respects the wireless path gain threshold \(G\).  
    
\end{enumerate}

A core feature of our novel SCoTT prompting paradigm is the explicit request for the model to explain its decisions within each subtask, specified in \textless Workflow\textgreater{} and \textless Reasoning Strategy\textgreater \ blocks. 
This leads to backtraceable outputs, e.g.:
\begin{quote}
    \emph{``I am choosing focus area 1—from (–1.5, 2.0) to (0.5, 3.0)—because the initial path runs through it and it contains predominantly blue (\textgreater0.8) path gain cells, whereas focus area 2 covers mostly orange (\textless0.5) regions that would risk falling below the average wireless path gain threshold."}
\end{quote}

SCoTT thus efficiently decomposes a single, complex path planning query into a sequence of focused CoT tasks, each scoped to a specific focus area and guided by an explicitly elicited reasoning strategy and workflow. 
This targeted breakdown prevents hallucinations and avoids context overflow, while producing concise and explainable outputs. 
We further note that the resulting ensemble of focus areas (represented as $k$d‐trees) from subtask 2 can serve as input to classical algorithms like DP‐WA* to reduce their search space.
We explore this neuro-symbolic extension in \refsec{sec:simulation_results}.
Furthermore, SCoTT uses the DT's path gain data intelligently, invoking it only during the fine‐grained refinement stage, thereby balancing data richness with prompt efficiency.
We demonstrate SCoTT's effectiveness across different VLM models in the next section.
\section{Simulation Results}
\label{sec:simulation_results}

\begin{table}[t]
    \centering
    \resizebox{0.5\textwidth}{!}{%
    \begin{tabular}{llccc}
        \toprule
        & & \makecell{\textbf{Path 1}: \\ Across the Room} & \makecell{\textbf{Path 2}: \\ Wall to Wall} & \makecell{\textbf{Path 3}: \\ Extreme Case} \\
        \midrule
        
        \multirow{4}{*}{\rotatebox[origin=tl]{90}{\makecell{\textbf{A*}}}} 
         & Avg. Path Gain & 0.34 & 0.06 & 0.10 \\
         & Path Length & 7.43 & 7.07 & 2.90 \\
         & Time & 2.23 & 2.24 & 1.01 \\
        \midrule
        
        \multirow{4}{*}{\rotatebox[origin=tl]{90}{\makecell{\textbf{N-WA*}}}} 
         & Avg. Path Gain & 0.46 & 0.31 & 0.22 \\
         & Path Length & 7.77 & 9.30 & 3.83 \\
         & Time & 4.30 & 5.32 & 3.53 \\
        \midrule
        
        \multirow{4}{*}{\rotatebox[origin=tl]{90}{\scriptsize \textbf{DP-WA*}}} 
         & Avg. Path Gain & 0.75 & 0.43 & 0.69 \\
         & Path Length & 9.15 & 10.21 & 9.63 \\
         & Time & 76.04 & 74.86 & 76.97 \vspace{0.1cm} \\
        \midrule
        
        \multirow{4}{*}{\rotatebox[origin=tl]{90}{\scriptsize \textbf{SCoTT}}} 
         & Avg. Path Gain & 0.72 & 0.41 & 0.62 \\
         & Path Length & 8.64 & 9.32 & 8.22 \\
         & Time & 56.50 & 52.20 & 59.10 \\
         \midrule
         \midrule
         
        \multirow{4}{*}{\rotatebox[origin=tl]{90}{\makecell{\scriptsize \textbf{SCoTT-} \\ \scriptsize \textbf{DP-WA*}}}} 
         & Avg. Path Gain & 0.73 & 0.43 & 0.66 \\
         & Path Length & 9.38 & 10.21 & 9.40 \\
         & Time & 30.01 & 29.17 & 33.76 \vspace{0.1cm} \\
        \bottomrule
    \end{tabular}
    }
    \caption{Comparison of path planning algorithms across three scenarios: Path 1 (Across the Room), Path 2 (Wall to Wall), and Path 3 (Extreme Case). Metrics include normalized average path gain, total path length (in meters), and runtime (in seconds). SCoTT results are reported for Llama-4-Scout-17B. All results are averaged across $N=10$ runs for fairness.}
    \label{tab:results}
    \vspace{-0.5cm}
\end{table}

\subsection{Experimental Setup}
\label{sec:experimental_setup}
We evaluate SCoTT across three challenging path planning examples, involving areas of both high and low wireless path gains. 
We conduct all simulations using the DT implementation from \cite{andrei2024digitaltwinningplatformintegrated}, where a 3D model of our lab at TUM was twinned into NVIDIA Omniverse \cite{nvidia_omniverse}. 
Therein, wireless ray tracing data was added using Remcom WirelessInsite \cite{remcom_wireless_insite} with accurate path gains for the twinned access point, which operates at 2.4~GHz using OFDM signaling with 1024 subcarriers, each spaced by 78.125~kHz.
All dimensions in the subsequent results are given in meters and path gains are normalized.
Whitespaces represent obstacles such as cupboards or tables. 
We leave the investigation of more complex DT environments to future work.
Our main interest is the validity of generated paths and corresponding performance metrics compared to classical A*, N-WA*, and DP-WA* approaches. 
For SCoTT, we use Llama-4-Scout-17B as our primary model and explore additional models in the further analysis.
We found that using five to seven focus areas ($N$) consistently produced the best outcomes.
In addition, in \refsec{sec:system_integration_and_practical_considerations}, we present corresponding ROS simulation results to complement SCoTT's path planning analysis with real-world navigation outcomes.

\subsection{Path Planning Results}
\label{sec:path_planning_results}
We investigate the following three path planning experiments, the results of which are shown in Table \ref{tab:results} (averaged for $N=10$ runs) for classical A*, N-WA*, DP-WA*, SCoTT, and the neuro-symbolic extension SCoTT-DP-WA*:

\subsubsection{Path 1 - Across the Room}

\begin{figure}[htbp]
    \centering
    \includegraphics[width=0.46\textwidth]{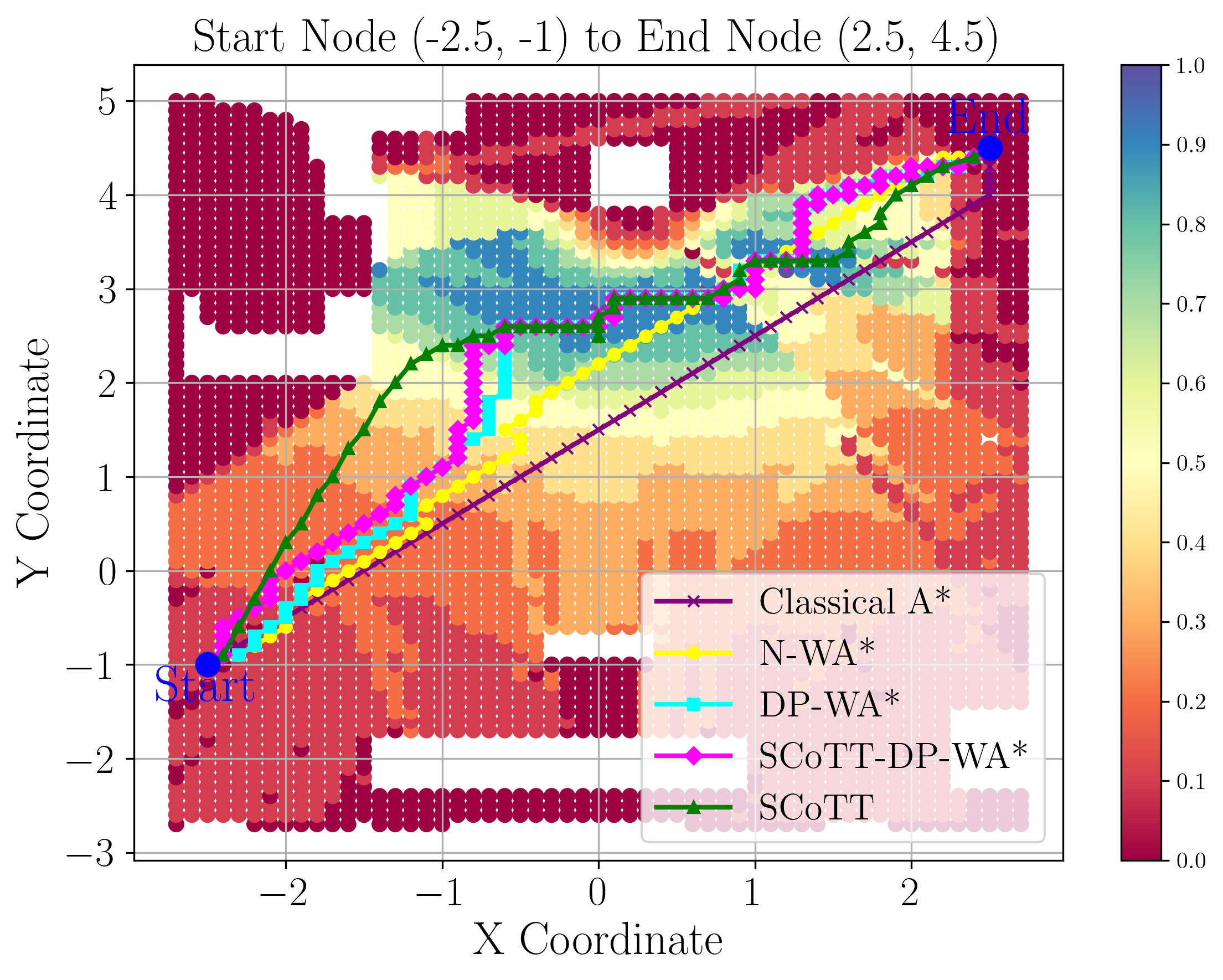}
    \caption{Results for path planning with $G$=0.7: SCoTT immediately moves toward the high gain blue area and overlaps with the optimal DP-WA* therein.}
    \label{fig:exp1}
\end{figure}

In \refig{fig:exp1}, a trajectory with average path gain threshold $G = 0.7$ is planned across the room from $n_{\text{start}} = (-2.5, -1)$ to $n_{\text{goal}} = (2.5, 4.5)$, traversing through areas with high, mid, and low path gains. 
The classical A* algorithm naturally chooses the shortest unobstructed path with length 7.43m, represented by the purple straight line, and thus has the lowest average path gain of 0.34. 
Similarly, N-WA* chooses a path with length 7.77m but deviates more toward higher gain areas, obtaining an average gain of 0.46. 
In contrast, the optimal DP-WA* path achieves the highest path gain of 0.75 at cost of a both increased path length at 9.15m and computation time of 76.04s, i.e., nearly 19 times higher than N-WA*. 
SCoTT, represented by the green path, is able to achieve a similar gain of 0.72 while slightly reducing the path length to 8.64m. 
The VLM-generated path in particular starts moving immediately to the higher gain areas and overlaps with DP-WA* therein before shortening its path toward $n_{\text{goal}}$ in the last few meters. 
Moreover, SCoTT produces the full path in 56s total, where it takes 5s to process the initial prompt and 7s, 15s, and 29s for executing subtasks 1, 2, and 3 (see \refig{fig:SCoTT_prompt}).
As noted in the previous section, DP-WA* can be significantly accelerated by reducing the search space using SCoTT's focus areas. 
This results in close performance to the original DP-WA* while reducing the computation time by more than 61\% to 30.01s, represented by the magenta SCoTT-DP-WA* path. 
This is especially useful because SCoTT can precompute focus areas and cache them for reuse in similar path-planning tasks.
Thus, only SCoTT or DP paths can successfully adhere to $G$ with SCoTT achieving very close performance to DP-WA*.

\subsubsection{Path 2 - Wall to Wall}

\begin{figure}[htbp]
    \centering
    \includegraphics[width=0.46\textwidth]{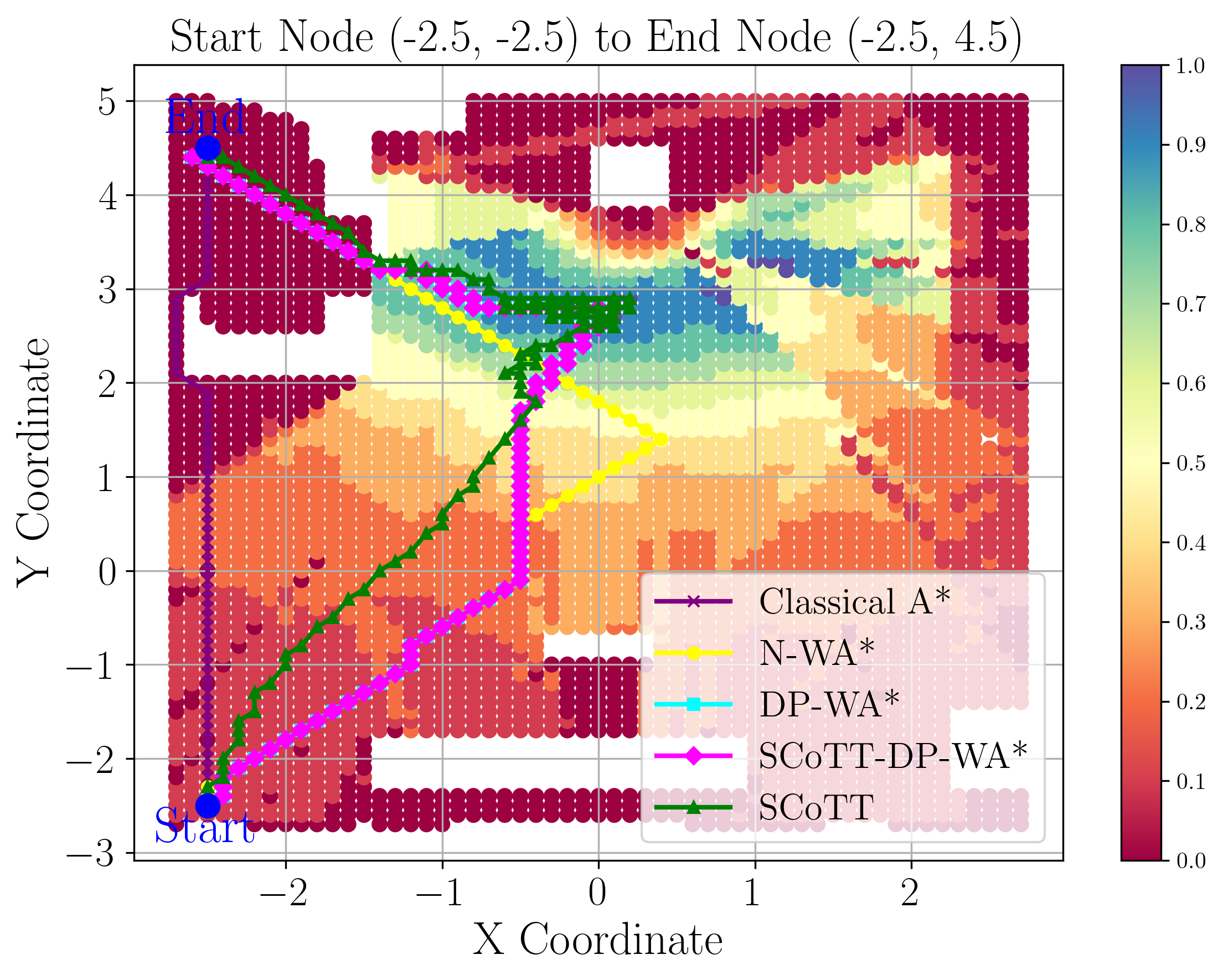}
    \caption{Results for path planning with $G$=0.4: All wireless-aware approaches, including N-WA*, avoid the shortest path and instead take a detour.}
    \label{fig:exp2}
\end{figure}

A more challenging scenario in a lower gain area is shown in \refig{fig:exp2}, where a trajectory with threshold $G = 0.4$ is requested. 
Intuitively, A* chooses the shortest path with length 7.07m and catastrophic gain 0.06, thus maintaining a straight line to the adjacent wall while going around the whitespace table obstacle. 
In contrast, all other paths, including N-WA*, take a detour into higher path gain areas to optimize for $G$. 
However, as before, only SCoTT and DP paths can adhere to $G$ with DP-WA* having the highest gain of 0.43, compared to SCoTT with 0.41.
In this example, SCoTT-DP-WA* is identical to DP-WA* whereas for \textit{Path 1} SCoTT's reduced focus areas may have omitted some few, but more optimal points, resulting in a slight difference compared to DP-WA*. 
As before, SCoTT tends to reduce the total path length while being faster than DP-WA* at around 52s. 
In addition, SCoTT-DP-WA* reduces DP-WA*'s computation time by more than 62\% to 29.17s.

\subsubsection{Path 3 - Extreme Case}

\begin{figure}[htbp]
    \centering
    \includegraphics[width=0.46\textwidth]{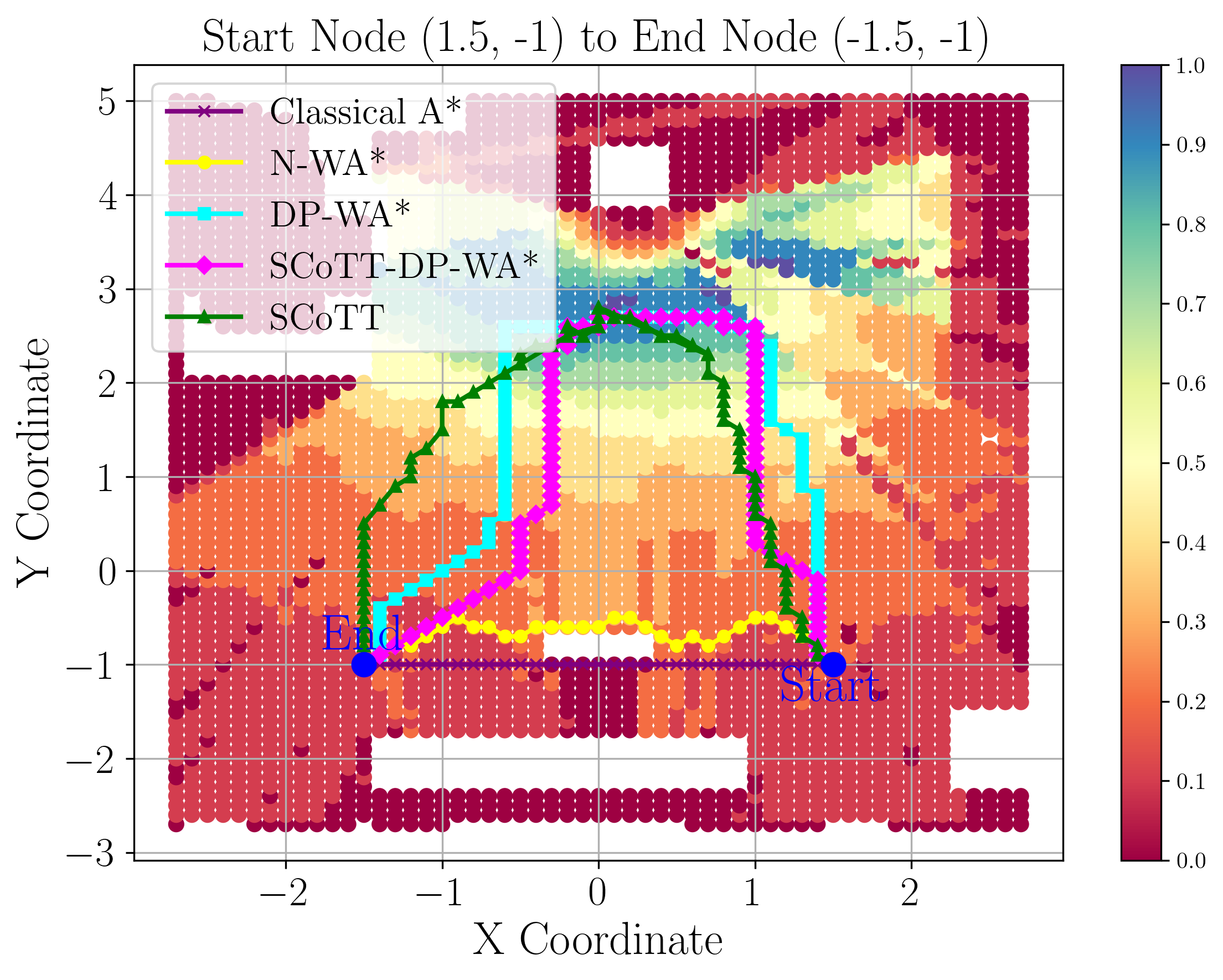}
    \caption{Results for path planning with $G$=0.6: N-WA* biases toward the shortest path while SCoTT and DP flavors effectively optimize for $G$.}
    \label{fig:exp3}
\end{figure}

Finally, we investigate an extreme case in \refig{fig:exp3}, where a trajectory with threshold $G = 0.6$ is requested for a very short path between $n_{\text{start}} = (1.5, -1)$ and $n_{\text{goal}} = (-1.5, -1)$. 
In this example, both A* and N-WA* maintain a short and roughly straight line from start to goal, where N-WA* biases toward the shortest path and thus cannot optimize the trade-off between optimal path gain and trajectory length. 
However, all SCoTT and DP paths take a semicircle-shaped detour into higher path gain areas with average gains of 0.69 for DP-WA*, 0.62 for SCoTT, and 0.66 for SCoTT-

\vspace{-0.5cm}

\vspace{-0.5cm}

\vspace{-0.1cm}

\newpage
\noindent
DP-WA*. 
Similarly, as before, SCoTT further reduces the total path length, while SCoTT-DP-WA* significantly reduces the computation time by 56\% from 76.97s to 33.76s, compared to DP-WA*. 
In this example, the threshold $G = 0.6$ is deliberately chosen to be higher in order to motivate such a stark detour from the shortest path, which can only be achieved by explicitly wireless-aware approaches such as DP-WA*, SCoTT, and SCoTT-DP-WA*.

In summary, SCoTT delivers results comparable to optimal algorithms such as DP-WA* while being competitive in time, typically completing about 20s faster.
Furthermore, SCoTT-DP-WA* accelerates DP-WA* by reducing the search space using SCoTT's focus areas, thus being up to 62\% faster in execution compared to DP-WA*. 
In our case study, the DT map comprised 3,854 waypoints, yet by leveraging SCoTT’s focus‐area reduction we cut the search space by approximately 48\%.
While evaluating SCoTT on larger DT maps would provide more insights into its scalability, such an extensive assessment falls outside the current scope of this exploratory work. 
We thus leave this investigation to future research.

\begin{table}[t]
    \centering
    \resizebox{0.5\textwidth}{!}{%
    \begin{tabular}{llccc}
        \toprule
        & & \makecell{\textbf{Path 1}: \\ Across the Room} & \makecell{\textbf{Path 2}: \\ Wall to Wall} & \makecell{\textbf{Path 3}: \\ Extreme Case} \\
        \midrule
        
        \multirow{5}{*}{\rotatebox[origin=tl]{90}{\makecell{\scriptsize \textbf{SCoTT} \\ \scriptsize \textbf{Llama-4}}}} 
         & Avg. Path Gain & 0.72 & 0.41 & 0.62 \\
         & Path Length & \textbf{8.64} & \textbf{9.32} & \textbf{8.22} \\
         & Time & \textbf{56.50} & \textbf{52.20} & 59.10 \\
         & Success Rate & 100 & 100 & 100 \\
        \midrule
        
        \multirow{5}{*}{\rotatebox[origin=tl]{90}{\makecell{\scriptsize \textbf{SCoTT} \\ \scriptsize \textbf{Llama-3}}}} 
         & Avg. Path Gain & 0.71 & 0.42 & 0.61 \\
         & Path Length & 8.79 & 9.41 & 8.39 \\
         & Time & 59.70 & 53.40 & \textbf{58.70} \\
         & Success Rate & 100 & 100 & 100 \\
        \midrule
        
        \multirow{5}{*}{\rotatebox[origin=tl]{90}{\makecell{\scriptsize \textbf{SCoTT} \\ \scriptsize \textbf{SmolVLM}}}} 
         & Avg. Path Gain & \textbf{0.78} & \textbf{0.49} & 0.67 \\
         & Path Length & 9.10 & 9.95 & 9.12 \\
         & Time & 61.10 & 59.80 & 62.10 \\
         & Success Rate & 100 & 90 & 90 \\
         \midrule
         
        \multirow{5}{*}{\rotatebox[origin=tl]{90}{\makecell{\scriptsize \textbf{SCoTT} \\ \scriptsize \textbf{Granite}}}} 
         & Avg. Path Gain & 0.75 & 0.46 & \textbf{0.68} \\
         & Path Length & 8.84 & 9.62 & 8.72 \\
         & Time & 60.10 & 58.40 & 63.40 \\
         & Success Ratio & 100 & 100 & 100 \\
        \bottomrule
    \end{tabular}
    }
    \caption{Comparison of SCoTT for different VLM backends across three scenarios: Path 1 (Across the Room), Path 2 (Wall to Wall), and Path 3 (Extreme Case). Metrics include normalized average path gain, total path length (in meters), runtime (in seconds), and success rate (in \%). Reported results are averaged across $N=10$ runs. Best results are highlighted in \textbf{bold}.}
    \label{tab:ablations}
    \vspace{-0.5cm}
\end{table}

\subsection{Model Variations}
\label{sec:model_variations}

We ablated SCoTT's generative model across four different open-source VLMs of varying sizes: Llama-4-Scout-17B \cite{llama4}, Llama-3.2-11B-Vision \cite{llama3}, SmolVLM \cite{smolvlm}, and Granite-Vision-3.2-2B \cite{granite}. 
All models were run locally on our compute cluster.
Table \ref{tab:ablations} reports runtime and performance metrics for each model, averaged over $N=10$ runs. 
We also report a success ratio, defined as the fraction of runs that produced valid paths, i.e., those respecting the average path gain constraints and successfully avoiding obstacles in the DT.

In general, the larger models, Llama-4 and Llama-3, achieve the best balance between average path gain and total distance, with Llama-4 outperforming Llama-3 in both speed and path length.
Surprisingly, the smaller models, SmolVLM and Granite, remain highly competitive.
However, these compact variants tend to favor higher gains at the expense of slightly longer routes, reflecting their bias toward blue- and green-rich regions during the visual reasoning step.
This behavior increases the overall runtime compared to the Llama models, as more waypoints are generated in areas with higher path gains, albeit only marginally.
In our experiments, nearly all models achieved a 100\% success ratio, except for SmolVLM, which occasionally failed on Paths 2 and 3 by missing obstacles. 
These results demonstrate that SCoTT’s prompting strategy is robust across different model architectures and that it can leverage lightweight models effectively at lower inference cost, which is particularly important for real-world deployments.
\section{System Integration and Practical Considerations}
\label{sec:system_integration_and_practical_considerations}

Ensuring that simulation-based plans translate seamlessly into real-world actions is essential for LLM-driven robotics.
In this section, we present an end-to-end pipeline that integrates SCoTT with our DT using ROS nodes, enabling direct robot control and navigation within the DT's environment.

\begin{figure}[htbp]
    \centering
    \includegraphics[width=1\linewidth]{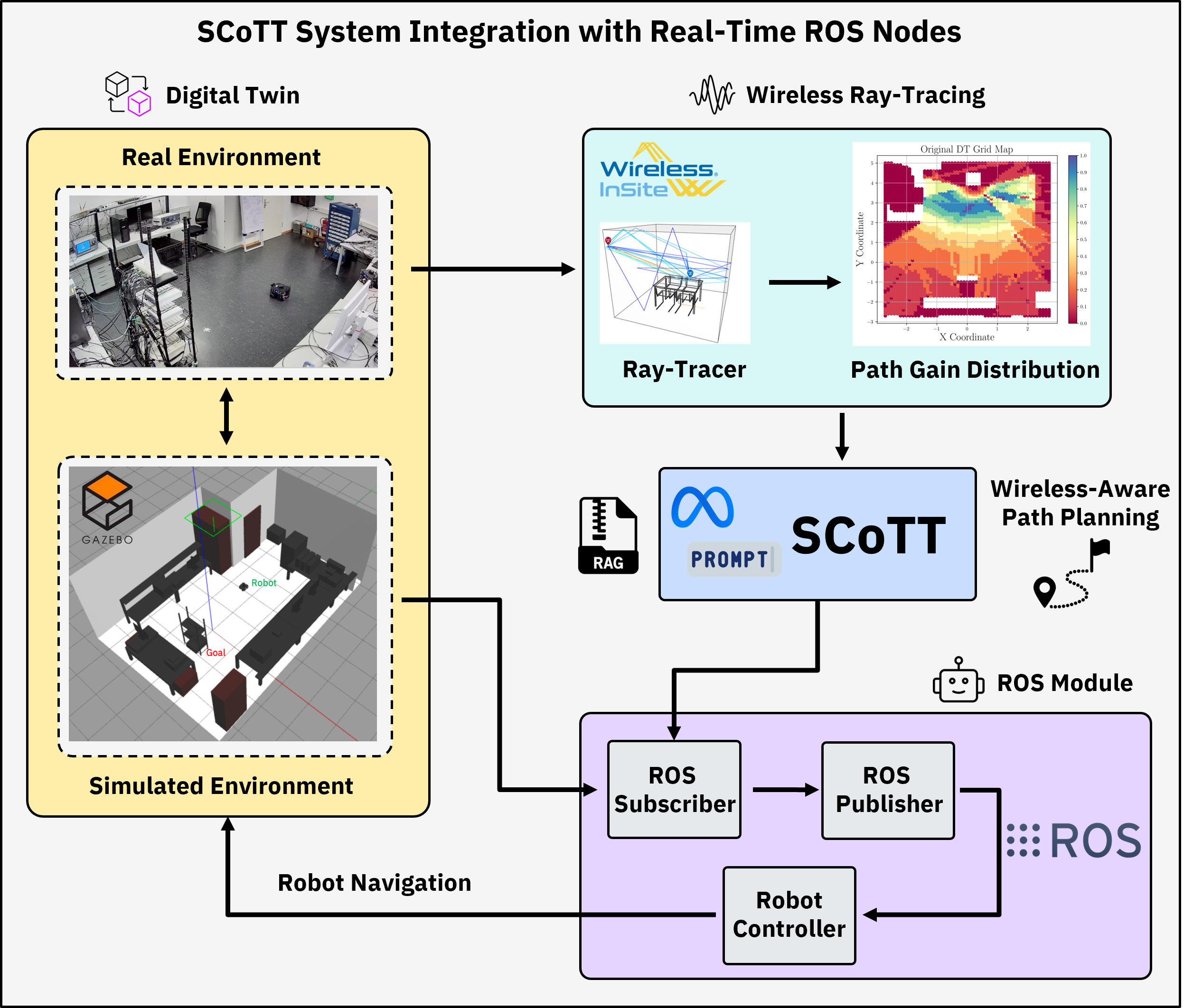}
    \caption{End-to-end integration of SCoTT with DTs and ROS. The DT supplies spatial information to a wireless ray tracer, which generates wireless heatmaps and path gain measurements. SCoTT then plans a wireless-aware path that is executed via ROS nodes within the DT environment.}
    \label{fig:system_integration}
\end{figure}

\subsection{System Integration with ROS}

\begin{figure}[t]
    \centering
    \includegraphics[width=0.9\linewidth]{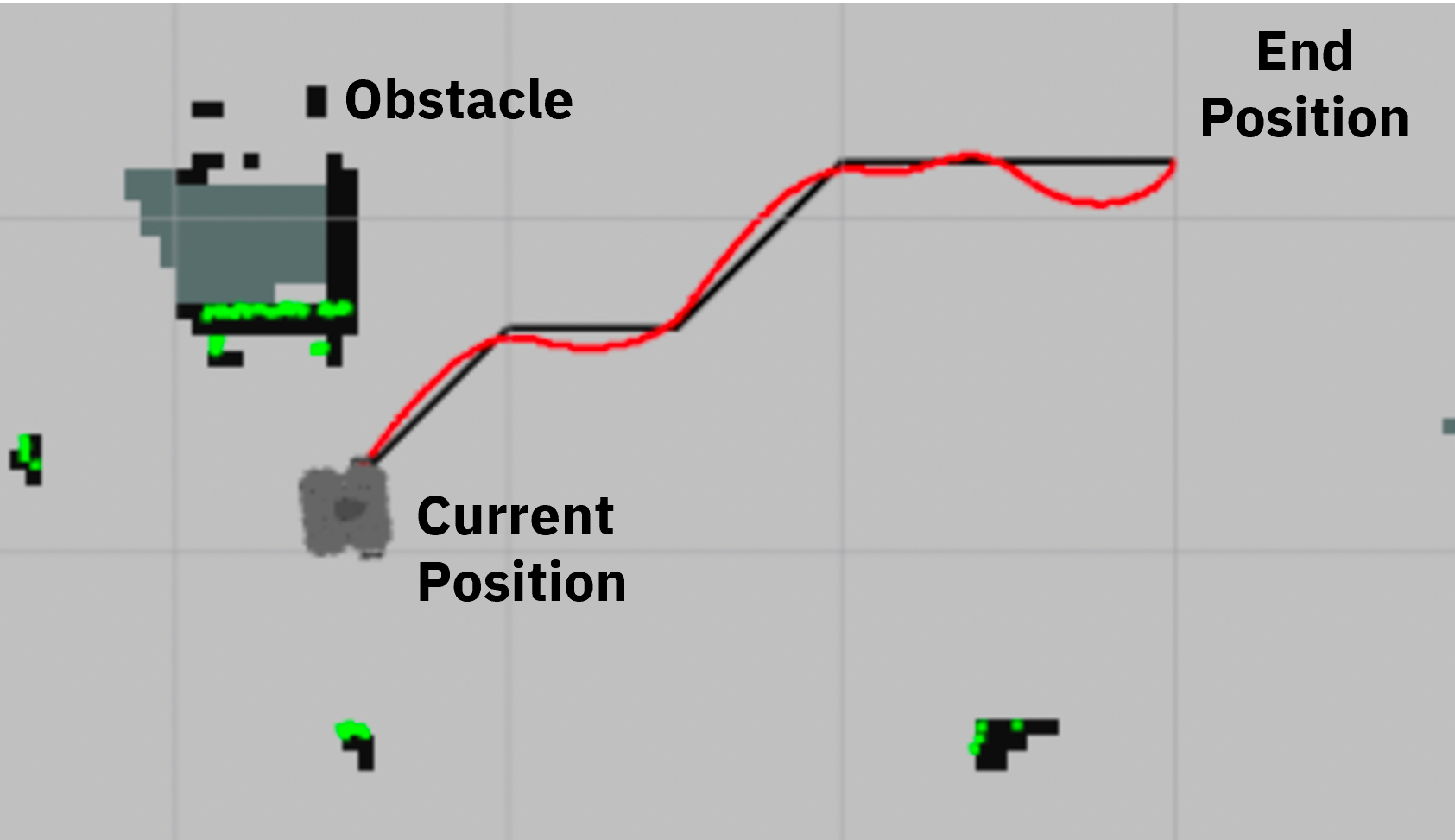}
    \caption{Snapshot of ROS-based navigation in the DT for \emph{Path 1 - Across the Room} (upper right hand corner): the SLAM-generated 2D occupancy grid shows SCoTT’s planned path (in black) and the robot’s executed trajectory (in red), with ROS nodes handling real-time execution and corrections. Deviations arise from odometry noise and minor localization shift.}
    \label{fig:ros_navigation}
    \vspace{-0.5cm}
\end{figure}

\refig{fig:system_integration} shows how SCoTT integrates with our DT via dedicated ROS modules.
In this setup, SCoTT plans a wireless-aware path using multi-modal data from the wireless ray tracer and spatial information provided by the DT.
The result is a sequence of 2D waypoints that guide the robot from its start to a goal position.
These waypoints feed into a ROS-based control loop for real-time navigation.
In general, navigation differs from path planning in that it covers the actual execution of the planned trajectory, including localization, motion control, and sensor feedback. 
A snapshot of this process is shown in \refig{fig:ros_navigation}, where the environment is rendered as a 2D occupancy grid map built with \emph{Simultaneous Localization and Mapping} (SLAM).
The red line traces the robot’s actual trajectory, closely matching SCoTT’s planned path in black.
Small deviations arise from real-time corrections as the robot compensates for odometry noise and minor localization drift, common in SLAM-based navigation.
We observe consistent results across all the examples from \refsec{sec:path_planning_results}.
This pipeline completes the full cycle: mapping the environment for obstacle avoidance, performing wireless-aware path planning with SCoTT, and executing the mission through the ROS control stack, ensuring realistic motion execution and map consistency.

\subsection{Practical Considerations}

SCoTT is fundamentally an inference‐only framework, such that no fine‐tuning or additional training is required. 
In addition, our experiments showed comparable performance using both large (Llama-4-Scout-17B) and compact (Granite-Vision-3.2-2B) models, suggesting that SCoTT can run on edge or near‐edge hardware with modest latency and at low inference cost. 
Moreover, SCoTT's data pipeline is mainly driven by the DT and wireless ray tracer, which generate the heatmap images and path gain JSON files. 
These assets can be fed into a lightweight RAG layer, which adds negligible overhead as data is handled on a per-need basis in each subtask. 
This model‐agnostic, modular design makes SCoTT both generalizable and practical for real‐time, 6G‐enabled robotic navigation.

\section{Conclusion}
\label{sec:conclusion}

In this paper, we proposed SCoTT, a wireless‐aware path‐planning framework that leverages multi‐modal VLMs to jointly optimize average wireless path gains and trajectory lengths.
By breaking a single planning query into a sequence of strategic CoT subtasks, each guided by an explicit reasoning strategy and workflow, SCoTT overcomes the hallucination and context‐window limitations of conventional language model-based planners and poses an alternative to prohibitive classical planners.
Through extensive simulations and ablation studies on three challenging scenarios and four different VLMs, we demonstrated that SCoTT achieves near‐optimal path gains within 2\% of our cost-optimal reference baseline (DP‐WA*).
We also showed that SCoTT can be used as a neuro-symbolic extension to classical algorithms, resulting in up to 62\% faster runtimes.
Furthermore, our model ablations showed that even compact models yield valid routes, making SCoTT suitable for on-device deployment at low inference cost.
Finally, we validated SCoTT’s practicality by integrating it end-to-end into a ROS-based navigation pipeline within a DT, demonstrating seamless integration and confirming its readiness for real-time, 6G-enabled robotic applications.

\hfill

\ifCLASSOPTIONcaptionsoff
  \newpage
\fi

\bibliographystyle{IEEEtran}
\bibliography{IEEEabrv,bibliography}

\end{document}